# From Feature Extraction to Classification: A Multidisciplinary Approach applied to Portuguese Granites

Vitorino Ramos, Pedro Pina and Fernando Muge

*Centro de Valorização de Recursos Minerais*
Instituto Superior Técnico
Av. Rovisco Pais, 1096 Lisboa Codex, PORTUGAL
email: {vitorino.ramos, ppina, muge}@alfa.ist.utl.pt

**Abstract:** The purpose of this paper is to present a complete methodology based on a multidisciplinary approach, that goes from the extraction of features till the classification of a set of different portuguese granites. The set of tools to extract the features that characterise polished surfaces of the granites is mainly based on mathematical morphology. The classification methodology is based on a genetic algorithm capable of search the input feature space used by the nearest neighbour rule classifier. Results show that is adequate to perform feature reduction and simultaneous improve the recognition rate. Moreover, the present methodology represents a robust strategy to understand the proper nature of the images treated, and their discriminant features.

## 1. INTRODUCTION

Ornamental stones are quantitatively characterised in many ways, mostly physical, namely, geological-petrographical and mineralogical composition, or by mechanical strength. However, the properties of such products differ not only in terms of type but also in terms of origin, and their variability can also be significant within a same deposit or quarry. Though useful, these methods do not fully solve the problem of classifying a product whose end-use makes appearance so critically important. Appearance is conditioned not only by the kind of stone but also depends on the subjective evaluation of beauty and hence of economic value, which are strongly influenced by supply and demand. Traditionally, the selection process is based on visual inspection, giving a subjective characterisation of the appearance of the materials. Thus, one suitable tool to characterise the appearance of these natural stones is digital image analysis. The identification, firstly, and the extraction, secondly, of the features that characterise a texture of a natural product are not anyhow easy tasks to accomplish. If the identification of the features (colour, size/shape, texture) that characterise a type of material may seem easier to list, the definition of a set of parameters to quantify those features becomes more problematic. Those parameters have to be the *good* ones, *i.e.*, they have not only to clearly characterise each feature for each type of material, but also, they should not be redundant. Moreover, image transforms applied to extract the features should be easily applied.

In what concerns classification, the n*earest neighbour rule* (k-NNR) was used. This technique is well known, is simple and has been considered to be a powerful classification technique [1], even for a small number of training prototypes. In that context, a method is described to implement nearest neighbour optimisation by g*enetic algorithms* (GA) (*i.e.* via feature reduction). In the basic nearest neighbour rule classifier, each training sample (described by their features) is used as a prototype and a test sample is assigned to the class of the closest prototype [2]. Its asymptotic classification error is bounded above by twice the Bayes error.

When the number of prototypes and/or their feature space is larger, this method requires a large memory space and long computing time. Because of this, the nearest neighbour rule has not found wide applications to solve pattern recognition problems. For example, the nearest neighbour classifier has equivalent recognition performance as a *radial basis function* (RBF) and *neural network* based classifiers [3]. Although it does not need any training to build a n*earest neighbour* classifier, it is most expensive to implement it in terms of memory storage and computing time.

Recently neural network based methods have been developed to overcome the implementation problem of nearest neighbour classifiers [4]. It has been shown that in some cases a small number of optimised prototypes can even provide a higher classification rate than all training samples used as prototypes.

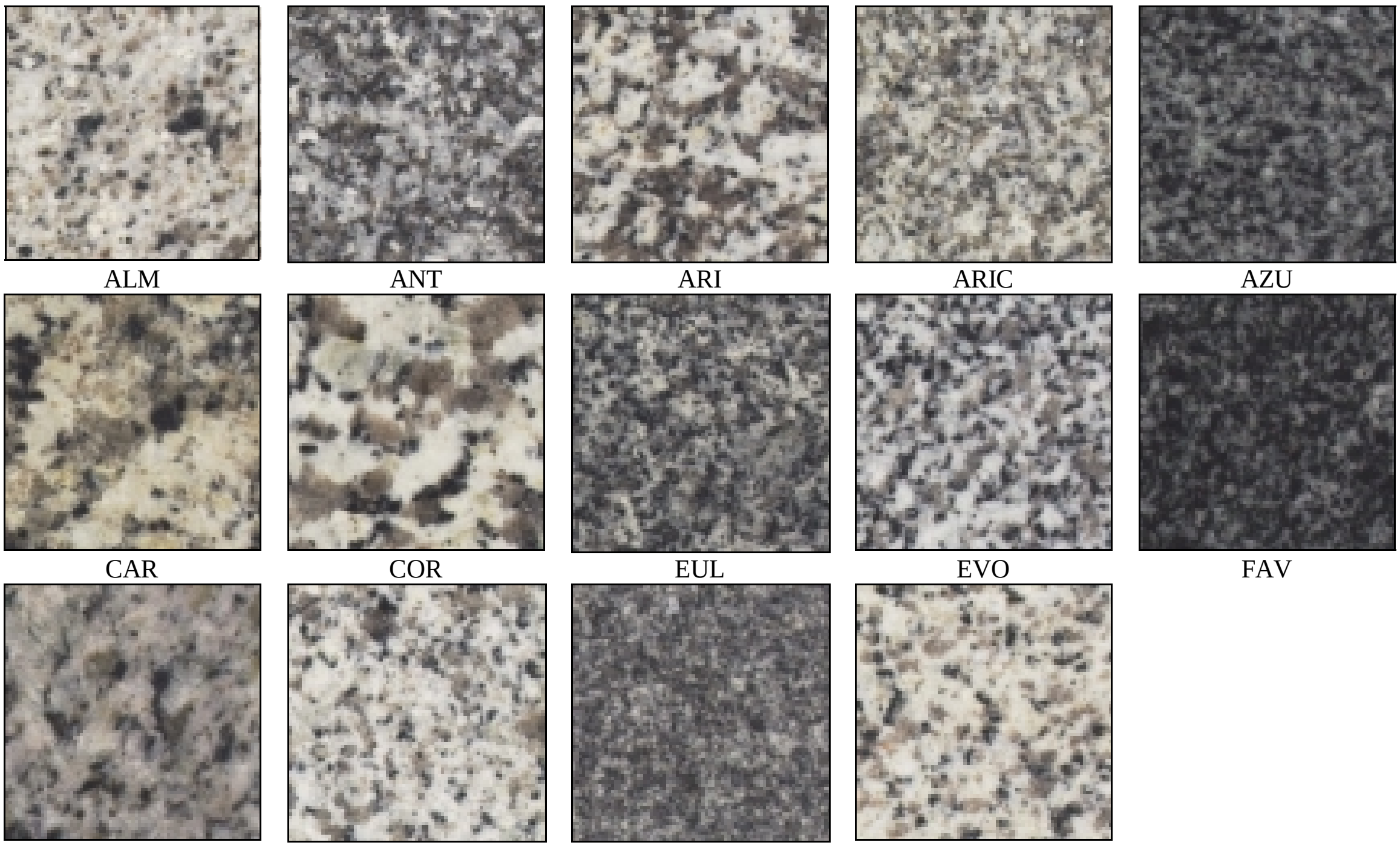

ALM ANT ARI ARIC AZU
CAR COR EUL EVO FAV
JAN SAL SPI VIM

Figure 1 – Portuguese grey granites, commercial designation
(each image represents an area of approximately 4 x 4 $cm^2$)

Another approach consists in using genetic algorithms based methods for searching the feature space to apply in nearest neighbour rule prototypes, which is the case presented in this paper. For instance, Brill *et al.* [5] used a k-NNR classifier to evaluate feature sets for counter propagation networks training. Some other authors used the same approach ([6],[7]) for another kind of classifiers.

## 2. GATHERING DATA

A collection of 14 Portuguese grey granites was previously defined to be studied [9]. Although this commercial label includes the real grey types, it also includes other similar colourless types (bluish, whitish and yellowish, for instance) (see figure 1 and table 1).

The samples of these types of granites used for the development and testing of our research are 30 cm x 30 cm polished tiles. The digital images were acquired (total set = 237 images) using a colour scanner with a predefined regulation set for the brightness and the contrast parameters [10] having a spatial resolution of 150 dpi and a spectral resolution of $256^3$ colours.

In the present case, a k-NNR classifier was applied to Portuguese granite classification. From 237 granite images, 50 were selected randomly for future performance evaluation (testing set). Each sample on the training set (187 samples) consists on a series of 117 features (Hue-Luminance-Saturation (HLS) histogram values and grey level granulometrical data - morphological openings of increasing size - see [9]), which were used on a first NNR classifier.

A GA was then built to search the entire space of combinations on these 117 features (exploring the N<117 features), using the classification rate of NNR classifier as its objective function (with Euclidean metric). We describe the classifier design and implementation procedures and we also verify the performance of the algorithm with experimental results. Results show that is possible to use this method also as a pre-processing tool for other kinds of classifiers.

| *Code* | | *Name* | *Samples* | | |
|---|---|---|---|---|---|
| | | | *ttl* | *trai* | *test* |
| 1 | ALM | Branco Almeida | 20 | 16 | 4 |
| 2 | ANT | Cinzento de Antas | 20 | 16 | 4 |
| 3 | ARI | Branco Ariz | 8 | 6 | 2 |
| 4 | ARIC | Cinza Ariz | 4 | 3 | 1 |
| 5 | AZU | Azulália | 20 | 16 | 4 |
| 6 | CAR | Branco Caravela | 20 | 16 | 4 |
| 7 | COR | Branco Coral | 20 | 16 | 4 |
| 8 | EUL | Cinzento Sta Eulália | 20 | 16 | 4 |
| 9 | EVO | Cinzento de Évora | 20 | 16 | 4 |
| 10 | FAV | Favaco | 15 | 11 | 4 |
| 11 | JAN | Jané | 20 | 16 | 4 |
| 12 | SAL | Pedras Salgadas | 10 | 7 | 3 |
| 13 | SPI | SPI | 20 | 16 | 4 |
| 14 | VIM | Branco Vimieiro | 20 | 16 | 4 |
| | | | 237 | 187 | 50 |

Table 1 – Types of Portuguese granites and number of samples used.

## 3. FEATURE EXTRACTION

Generically, the extraction of features by means of image analysis and mathematical morphology techniques is implemented in 2 stages: a global and a local analysis. It consists on the extraction of features before (global analysis) and after (local analysis) the segmentation or *phase* classification procedures. This approach is general and can be applied not only to characterise slab surfaces of granites but also other types of ornamental stones.

### 3.1. The Global Analysis Stage

The g*lobal analysis stage* consists on the first attempt to characterise the textures, and it is applied to the colour images before the segmentation or *phase* classification procedure. The main idea, which is also the main expectation, at this phase is that the features extracted could be sufficient to discriminate several types of materials, being not needed to proceed to a local analysis stage. The features that characterise the different structures concern the colour, the size and the texture of their components.

- *Colour features*

Due to the fact that this set of 14 granites is of the colourless type (the colour histograms are mainly frequented around the main diagonal of the RGB cube), a simplification of the three colour images into one grey level image is simple to obtain and is correct: the R, G and B components for each sample of all types of granites present a very similar behaviour and the respective average images (intensity) do not differ very much from each individual one. Also, HLS components were individually considered.

- *Size and shape features*

In what concerns the size of the mineralogical components of the granites a size distribution can be directly achieved through the grey level images. The size distribution by grey level granulometry can be computed by morphological openings or closings of increasing size. These transforms are simultaneously applied to all *phases* present in the image.

The opening (erosion followed by dilation) and the closing (dilation followed by erosion) have granulometric properties [11] once are increasing, extensive (closing) and anti-extensive (opening) and idempotent [12]. In figures 2 and 3, are presented the different opened images with a hexagon of size 3, 5 and 10 for the types AZU, EUL

The grey level granulometry consists on measuring the volume *V* of the function *f(x)* after the application of the opening, $\gamma^{B(r)}\left[f(x)\right]$, or closing, $\varphi^{B(r)}\left[f(x)\right]$, with the structuring element *B* of size *r*. The respective cumulative distribution functions $G_f(r)$ are, for the openings, given by

$$G_f(r) = \frac{V\left[f(x)\right] - V\left[\gamma^{B(r)}(f(x))\right]}{V\left[f(x)\right]} \qquad (1)$$

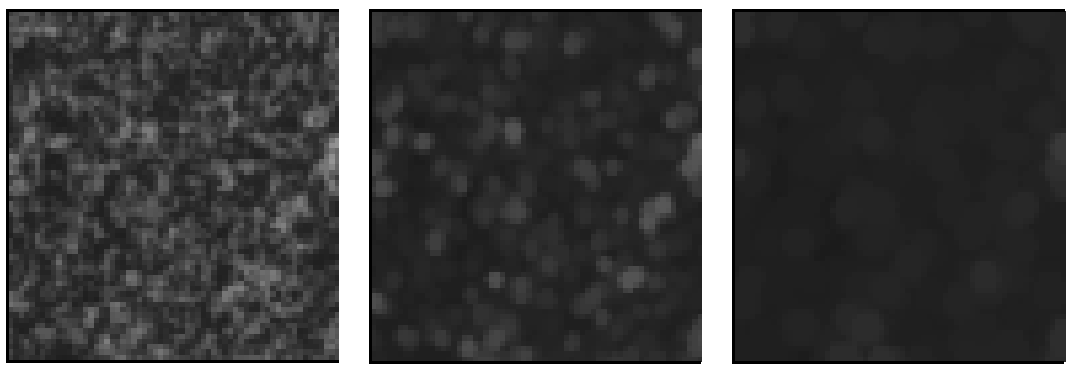

(a) (b) (c)

Figure 2 – Hexagonal openings of AZU granite of size: (a) 3; (b) 5 and (c) 10.

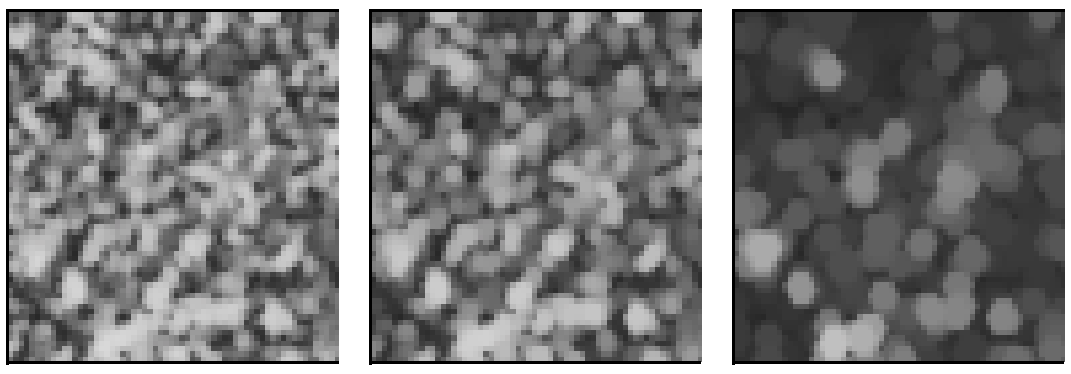

(a) (b) (c)

Figure 3 – Hexagonal openings of EUL granite of size: (a) 3; (b) 5 and (c) 10.

A similar grey level granulometry can be computed by applying closings of increasing size. The corresponding granulometrical curves reflect the size distribution of their darker components.

Besides the information related to the size given by the classical grey level granulometry (measure of the volume), another measure can be achieved by associating it the distribution of the grey levels. This measure recently introduced [13], consists on the computation of the grey level distribution for opened or closed images of increasing size. The resulting diagram is of bidimensional type and incorporates this way both size and intensity or grey level information.

This diagram is built using a family of openings or closings of increasing size using the cylindrical structuring element *B(r,k)* of radius *r* and height *k*.

The size-intensity diagram by openings of a function *f*, *SI(f)*, can be defined as:

$$SI_f(r,k) = \mathrm{A}(\gamma^{B(r,k)}(f)) \qquad (3)$$

where $\mathrm{A}(f)$ is a measuring function of the number of non-zero pixels of *f*:

$$\mathrm{A}(f) = Meas\left\{x \in Z^2 : f(x) \neq 0\right\} \qquad (3)$$

The size-intensity diagram has granulometrical properties because satisfies Matheron axioms extended to two parameters.

For the opening of *r=0*, the column $SI_f(0,k)$ of the size-intensity diagram corresponds to the grey level distribution of the initial image. It can also be shown that each row (fixed *k*, varying *r*) of the size-intensity diagram gives the granulometry of the binary image thresholded at level *k* [13].

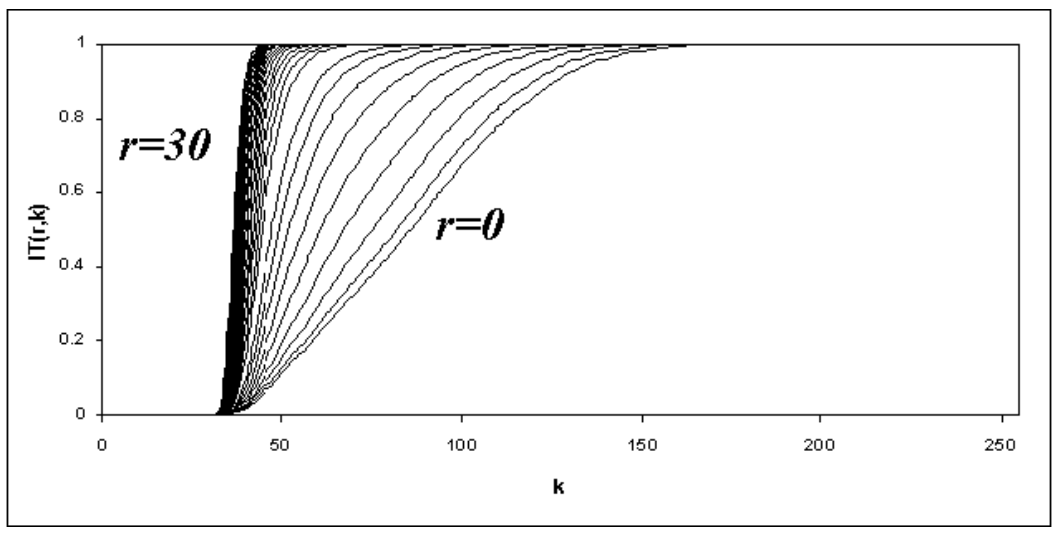

Figure 4 – Size-intensity diagram by openings of increasing size for AZU granite

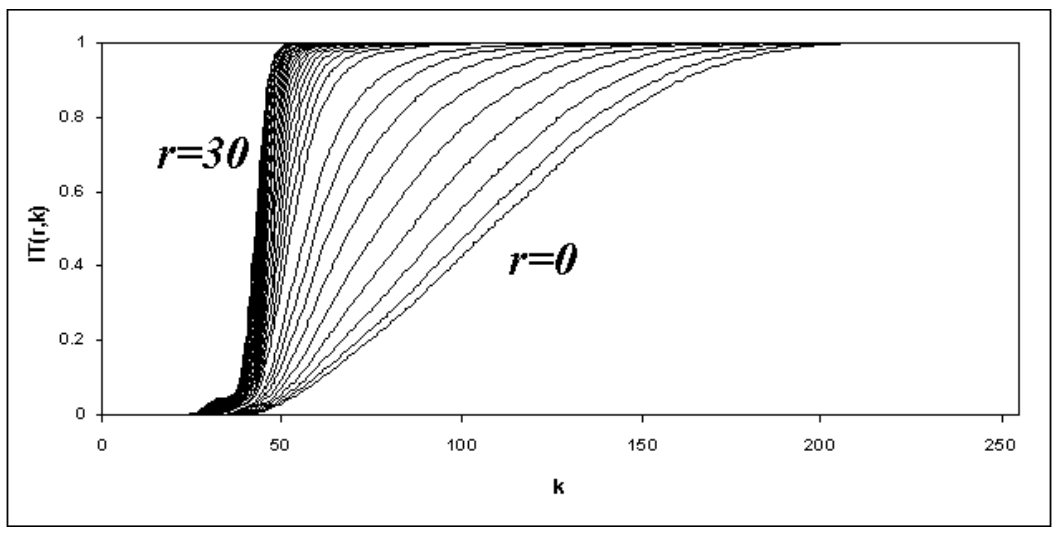

Figure 5 – Size-intensity diagram by openings of increasing size for EUL granite

In figures 4 and 5 are presented the size-intensity diagrams for AZU and EUL types, respectively. Differences on the diagrams constructed with curves from *r* varying from 0 to 30 can be noticed: besides the similar size of their components, EUL type is constituted by structures that are lighter than the ones of AZU type.

### 3.2. The Local Analysis Stage

At the *local analysis stage* all phases in the images must be segmented or classified [14]. This stage consists on the individual study of each classified phase and can be seen as a more detailed evaluation of the structures. At this stage, the features considered are the size, shape and spatial arrangements of each phase taken individually. In the present study it was not necessary to proceed to a local analysis.

## 4. The Training and Testing Set

From the features extracted in each Portuguese granite image, two training/testing sets were build: RGB with 27 features (mainly RGB histogram values) and LOT with 117 features (HLS histogram values and grey level granulometry data - morphological openings of increasing size). On the other hand, the test set was composed with 50 random and independent images representing the 14 different types of Portuguese Granites (see fig.1 - for each sample, commercial designations were used). This way, training set RGB is represented by a matrix of 187 lines (187 images representing 14 different types of Portuguese granites) and 27 columns (27 features). Testing set RGB was represented by a matrix of 50 lines (50 testing images representing also the 14 granites) and 27 columns. Similarly for LOT, *i.e.*, the training set with a matrix of 187 lines by 117 columns and the testing set by a matrix of 50 lines and 117 columns. As said before, feature extraction was mainly done with grey level intensity parameters, for each image.

## 5. The *K*-Nearest Neighbour Rule (*k*-NNR).

Nearest neighbour methods are among the most popular for classification [10]. They represent the earliest general (non-parametric) methods proposed for this problem and were deeply investigated in the fields of statistics and (specially) pattern recognition. Recently renewed interest on them emerged in the connectionist literature ("memory" methods) and also in machine learning ("instance-based" methods). Despite their basic simplicity and the fact that many more sophisticated alternative techniques have been developed since their introduction, nearest neighbour methods still remain among the most successful for many classification problems.

The *k*-NNR assigns an object of unknown class to the plurality class among the *k* labelled "training" objects that are closer to it. Closeness is usually defined in terms of a metric distance on the Euclidean space with the input measurement variables as axes. Nearest neighbour methods can easily be expressed mathematically. Let $x$ be the feature vector for the unknown input, and let $m_1$, $m_2$, …, $m_c$, be the templates (*i.e.*, perfect, noise-free feature vectors) for the $c$ classes. Then the error in matching $x$ against $m_i$, is given by:

$$\|x - m_i\| \quad (4)$$

Here $\|u\|$ is called the norm of the vector $u$. A minimum-error classifier computes $\|x\text{-}m_i\|$ for $i = 1,\ldots c$ and chooses the class for which this error is minimum. Since $\|x\text{-}m_i\|$ is also the distance from x to $m_i$, we call this a minimum-distance classifier. Naturally, the $n$ dimension Euclidean distance $d$, of $x$ ($n$ dimensional feature vector) to a training sample $m$, is:

$$d = \sqrt{\sum_{i=1}^{i=N} \left(x_i - m_i\right)^2} \quad (5)$$

The previous approach may be extended to the *k-nearest neighbour* rule (k-NNR), where we examine the labels on the *k*-nearest samples in the input space and then classify them by using a voting scheme. Often in $c$=2 class problems, $k$ is chosen to be an odd number, to avoid *ties*. Other significant concerns and possible extensions include the use of a rejection option in instances where there is no clear *winner*, and the finite sample size performance of the *NNR*.

| | *1-NNR RGB* | *1-NNR LOT* | *3-NNR RGB* | | | | *3-NNR LOT* | | | |
|---|---|---|---|---|---|---|---|---|---|---|
| *Recognition Rate* | 96 % | 98 % | 94 % | | | | 98 % | | | |
| *Number of Errors (% in 50)* | 2 = 4 % | 1 = 2 % | 3 = 6 % | | | | 1 = 2 % | | | |
| SAMPLE | Dec. | Dec. | 1ºN. | 2ºN. | 3ºN. | Dec. | 1ºN. | 2ºN. | 3ºN. | Dec. |
| ARI1-2 | Cor 5-4 | | Cor 5-4 | Vim 2-4 | Vim2-3 | Vim | | | | |
| ARIC1-2 | | | Aric 1-4 | Aric 1-3 | Car 4-4 | Aric | | | | |
| EUL1-3 | | | | | | | Eul 1-1 | Eul 1-4 | Ant 2-4 | Eul |
| SAL3-1 | | Vim 3-1 | Vim 3-1 | Sal 4-1 | Sal 1-1 | Sal | Vim 3-3 | Sal 1-1 | Sal 3-2 | Sal |
| SAL5-1 | Vim 3-1 | | Vim 3-1 | Vim 4-1 | Vim 3-3 | Vim | | | | |
| VIM3-2 | | | | | | | Vim 3-4 | Sal 5-2 | Sal 3-2 | Sal |
| VIM4-3 | | | Vim 1-3 | Sal 5-2 | Sal 4-2 | Sal | | | | |
| VIM5-1 | | | Vim 4-4 | Vim 5-4 | Sal 2-1 | Vim | | | | |

Table 2 - Portuguese granite classification - testing samples for which occurred an recognition error (black cells) or where the voting scheme 3-*NNR* was not in agree. Note the type of *confusions:* ARI classified as COR (1-*NNR* and 3-*NNR* / RGB), SAL as VIM (1-*NNR* and 3-*NNR* / RGB ; 1-*NNR* / LOT), and VIM as SAL (3-*NNR* / RGB and LOT) (Results based on a random and independent testing set with 50 samples).

Given a vector *x* and a training set *H*, whose cardinality may be large, the computational expense of finding the nearest neighbour of *x* may be significant. For this reason, frequent attention has been given to efficient algorithms. The computational savings are typically achieved by a pre-ordering of samples in *H*, combined with efficient (often hierarchical) search strategies (for extended analysis see [15]).

## 6. GRANITES CLASSIFICATION

In order to compare future results, 1-*NNR* and 3-*NNR* were implemented using RGB and LOT training sets for Portuguese granite classification (*i.e.*, with all features: 27 in RGB; 117 in LOT). Extended results are presented on table 2. The best results in terms of successful recognition were obtained using LOT training set (with 117 features) and 1-NNR (98%). In global terms, recognition errors occur with VIM and SAL samples.

To get out an idea how these training samples are arranged in their respective *n* dimensional spaces, *principal component analysis* (PCA) was applied. PCA is perhaps the oldest and best-known technique in multivariate data analysis and data compression [16]. In the present case, PCA has been used to obtain the first two principal components (PCs) (*i.e.*, the PCs corresponding to the first few dominant eigenvalues of the covariance matrix of the stationary input patterns) and this PCs were used subsequently as 2D *feature* scatterplot axis (figure 6 shows the scatterplot for the RGB training set; figure 7 for the LOT training set).

## 7. FEATURE SPACE REDUCTION VIA GENETIC ALGORITHMS

In order to reduce the input feature space, and hypothetically improve the recognition rate, genetic algorithms were implemented. The idea is to reduce the number of features necessary to obtain at least the same recognition rates. *Genetic algorithms* (GAs - [16], [10] and [17]) are search procedures based on the mechanics of natural selection and natural genetics. The GA were developed by *John H. Holland* in the 1960's to allow computers to evolve solutions to difficult search and combinatorial problems, such as function optimisation and machine learning. The basic operation of a GA is conceptually simple (canonical GA):

(1) To maintain a population of solutions to a problem,
(2) To select the better solutions for recombination with each other, and
(3) To use their offspring to replace poorer solutions.

The combination of selection pressure and innovation (through *crossover* and *mutation* - genetic operators) generally leads to improved solutions, often the best found to date by any method (see [12], [17]). For further details on the genetic operators, GA codification and GA implementation, the reader should report to [18].

Usually each individual (*chromosome*; pseudo-solution) in the population (say 50 individuals) is represented by a binary string of 0's and 1's, coding the problem which we aim to solve. Here, the aim is to analyse the combinatorial feature space impact on the recognition rate of n*earest neighbour* classifiers. In other words, the GA will explore the $N$<117 features of LOT training set ($N$<27 in the RGB case) and their combinations. An efficient genetic coding for a feature sub-space is then represented by each GA individual - i.e. an hypothetical classification solution, via a reduced group of features. The GA fitness function is then given by the 1-*NNR* classifier recognition rate and by the number of features used on that specific NNR.

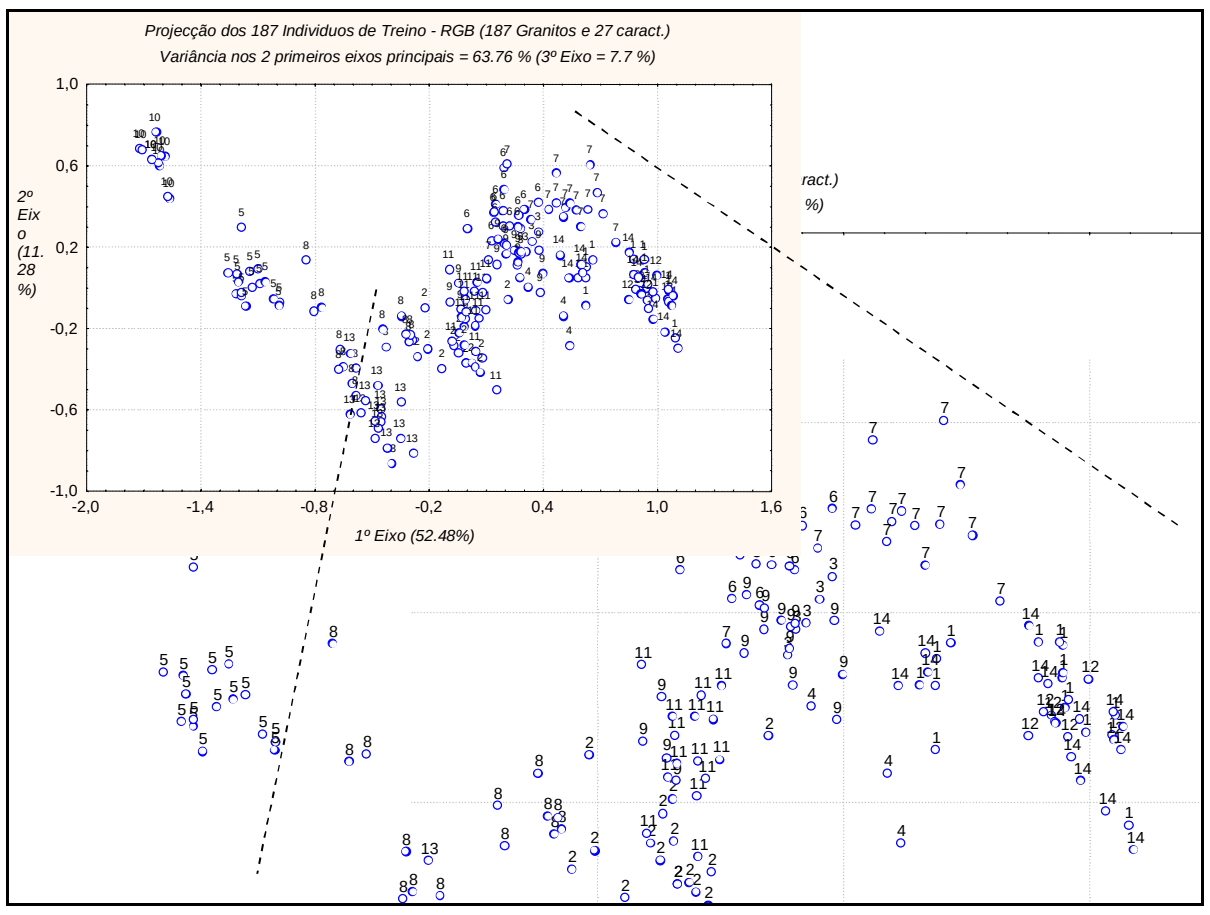

Figure 6 - PCA projection (187 Portuguese granite images) from RGB training set (with 27 features for each sample).

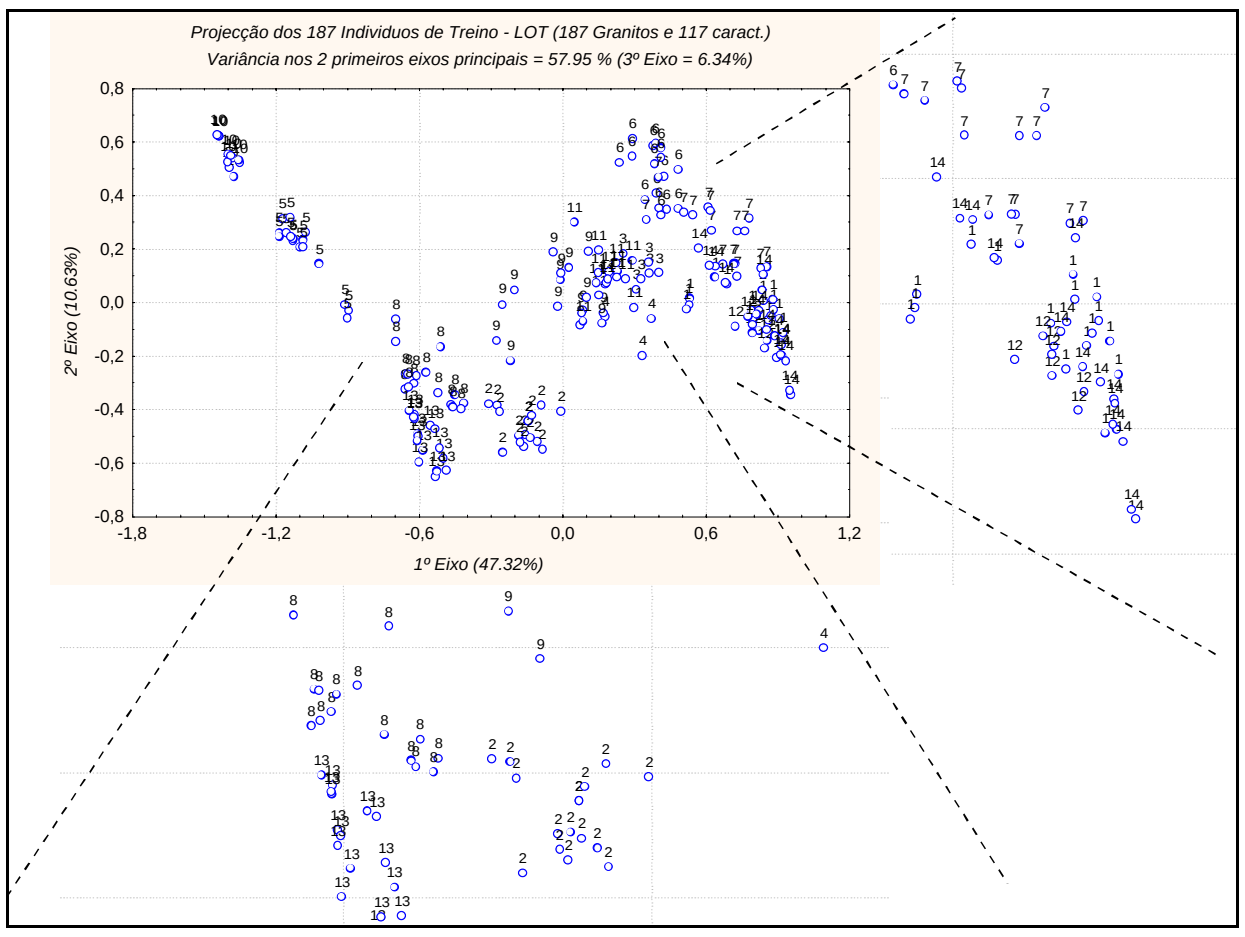

Figure 7-PCA projection (187 images) from LOT training set (with 117 features for each sample).

In that way, for each GA generation and for each individual (pseudo-solution) is then performed the nearest neighbour classification. The results are then used on the GA again. The algorithm proceeds their search until a stopping criterion is achieved. After convergence, the final solution points out what features among 117 can maximise the fitness function. The overall GA search space is given by the sum of all combinations of 116 features in 117, plus the 115 features in 117, …, plus all combinations of 2 features in 117. For the $i$ individual the fitness function can be expressed as:

$$fit[i] = \alpha\, hits[i] - \beta\, nf[i] \qquad (6)$$

where $\alpha$ and $\beta$ are real valued constants ($\alpha + \beta = 1$), and $hits[i]$ represents the number of images well recognised among the testing set, and $nf[i]$ the number of features used on *NNR* classification. The representation (GA coding) for each solution is then achieved by means of a binary string 117 bits long (for the LOT training set case) and 27 bits long (for RGB training set case), *i.e.*, if the $n^{th}$ bit is 1, then the $n^{th}$ feature is used on the *NNR* classification; if not (= 0) that specific feature is not used on the classification. Similar coding strategies were implemented by several authors, yet for different classification purposes [18].

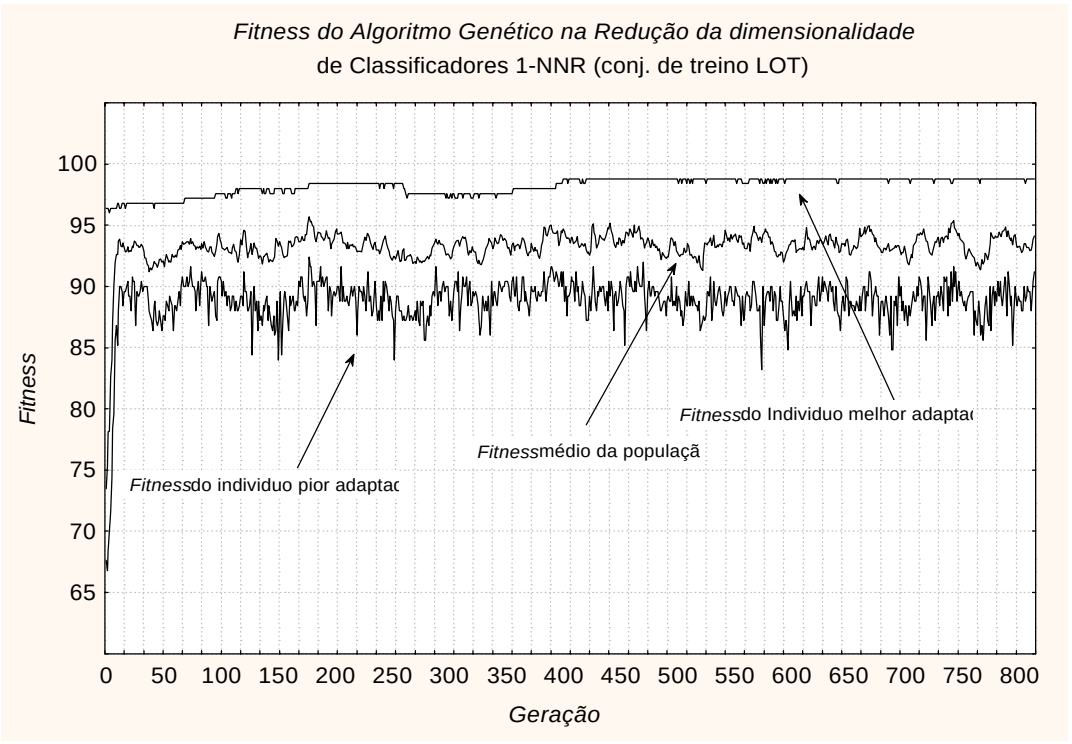

(a)

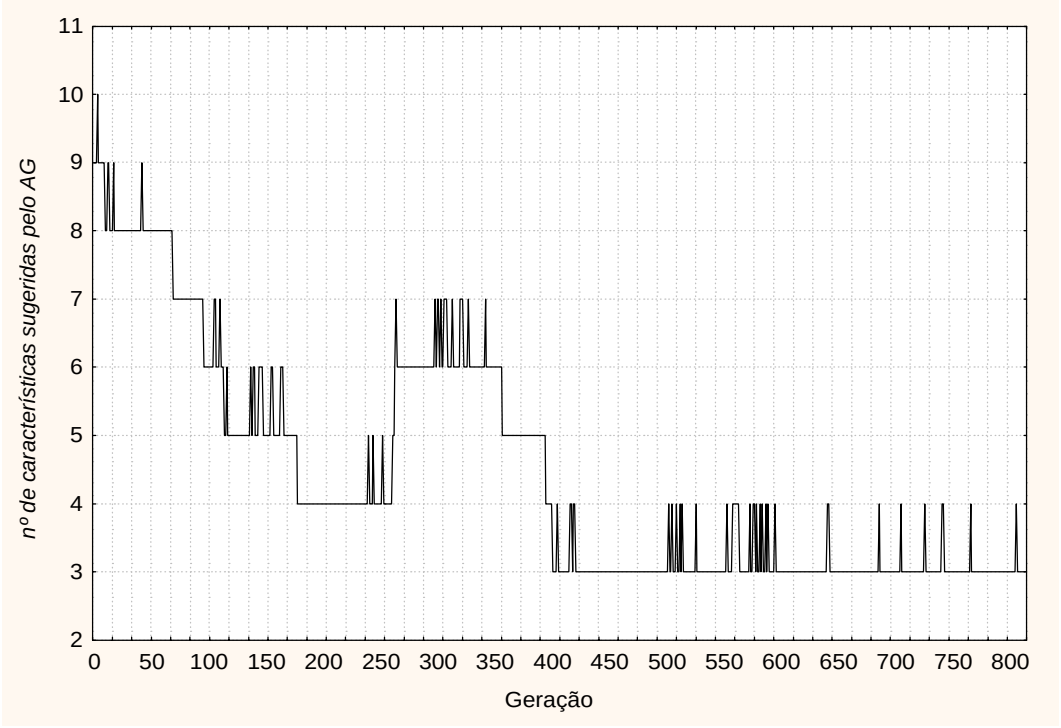

(b)

Figure 8 - (a) Minimum, best and median fitness on the 50 GA individual (training set LOT), (b) number of features in each generation, for the same GA (Portuguese granite classification via feature reduction on NNR GA).

Table 3 resumes the GA parameters used for feature reduction and classification by means of this hybrid method. Figure 8 (a,b) shows how convergence was achieved, generation after generation, in terms of population fitness (best individual, poor individual, and median population fitness), as well as in terms of feature reduction. Figure 9 gives an idea of the discrimination degree achieved in graphic mode (results for the LOT training set). Once the GA final solution for this set is constituted by 3 features, it is possible to build 3 scatterplots (each one with 2 features). Finally, figure 9 also presents 3 zooms from one scatterplot, allowing to see in detail each cluster (granite type).

| | *(1)* | *(2)* |
|---|---|---|
| Nº of original features | 117 | 27 |
| *bits* for each individual | 117 | 27 |
| individuals for each generation | 50 | 216 |
| Generations run | 814 | 250 |
| Time (hours) | 2.37 | 1.31 |
| Recognition rate on the Test set (50 images) | 100 % | 100 % |
| Crossover probability | 1.00 | 1.00 |
| Mutation probability | 0.90 | 0.90 |
| final features | 3 | 5 |
| Features suggested by the GA - # | 70, 101, 112 | 2, 9, 10, 17, 21 |
| $\alpha$ | 0.6 | 0.4 |
| $\beta$ | 0.6 | 0.4 |
| Random Seed | 12957 | 1547 |

Table 3 - GA parameters used on feature space reduction for classification via n*earest neighbour* rule (Computer: PENTIUM 166 MHz / 32 Mb RAM).

## 8. CONCLUSIONS

In this paper are only presented the tools chosen to extract features from polished slab surfaces of Portuguese grey granites. They completely cover the features related to the colour, the size and the spatial arrangement of the structures that constitute the granites, and are not very difficult to implement and to apply. Results show that this hybrid strategy is highly promising. In fact, not only recognition rate was improved, as the number of features necessary for successful Portuguese granite image classification was substantially reduced. Using LOT training set, for instance, was possible to improve the recognition rate by 2 % (achieving 100%) and reducing the number of important features from 117 to 3 (representing a reduction rate of 97 %). Computation time was in the last case 142 minutes, but can be reduced using new types of algorithms, strictly involved on the computation of nearest neighbours (as said before, typically achieved by a preordering of samples in the input space). However, this time is spent only once, *i.e.*, on the training phase. After that, its possible to classify images based on a reduced number of features, improving real time computer calculus on nearest neighbour relations, since the dimension of feature space was significantly reduced. Size-intensity diagram revealed to be appropriated on granite textures, since its main characteristics are related to these kind of features. On this context, this methodology points out the sub-features that are really important for successful discrimination. This way, it is possible to understand rigorously and to improve the knowledge on the proper nature of the digital images we are working with.

*ACKNOWLEDGEMENTS*

Some of the methodologies and some of the results presented were developed and obtained in the frame of the project *COSS - Characterisation of Ornamental Stones Standards by Image Analysis of Slab Surface* (contract no SMT4-CT95-2028) funded by the European Union/DGXII under its *Standards, Measurements and Testing* programme.

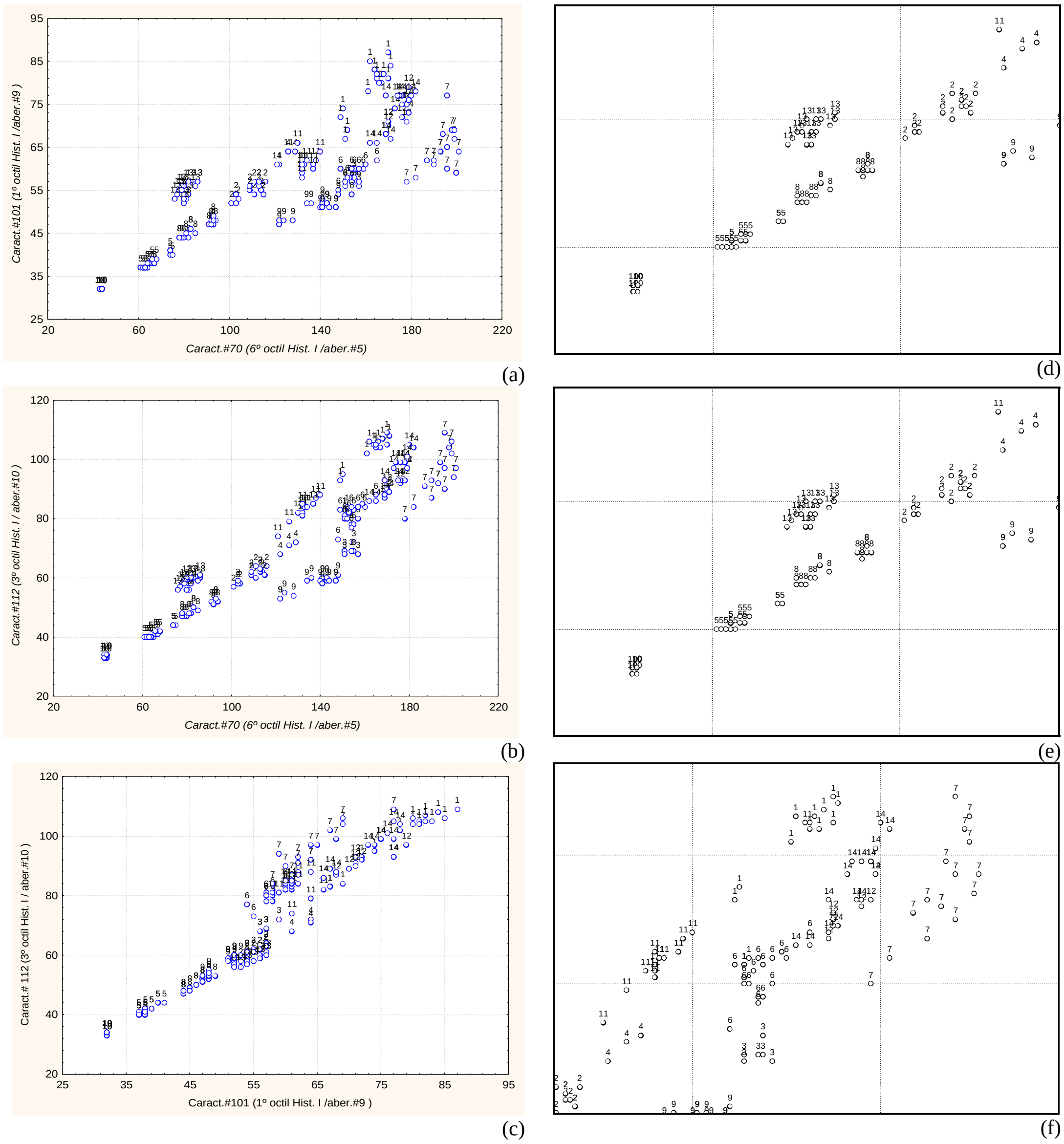

Figure 9 – Scatterplots (a,b,c) resuming the 3D mapping of 187 Portuguese granite images (LOT training set), via the 3 features (in 117) suggested by the genetic algorithm. Sample codes for each type of Granite are the same as in figure 4. Zooms (d,e,f) on the previous scatterplot (figure 7-b) - Features #70 and #112.